\pdfoutput=1

\documentclass[conference]{IEEEtran}

\usepackage{cite}
\usepackage{amsmath,amssymb,amsfonts}
\usepackage{graphicx}
\usepackage{textcomp}
\usepackage{bm}
\usepackage{booktabs}
\usepackage{array}
\usepackage[final]{microtype}
\newcolumntype{L}[1]{>{\raggedright\arraybackslash}p{#1}}
\newcolumntype{C}[1]{>{\centering\arraybackslash}p{#1}}
\usepackage{algorithm}
\usepackage{algpseudocode}
\usepackage{hyperref}

\def\BibTeX{{\rm B\kern-.05em{\sc i\kern-.025em b}\kern-.08em
    T\kern-.1667em\lower.7ex\hbox{E}\kern-.125emX}}

\begin{document}

\title{Contextualized Early Detection of Online Firestorms:\\
A Sequential LLM-Based Approach}

\author{%
  \IEEEauthorblockN{Besim Shala, Peter Mandl, Andreas Humpe, and Martin H\"{a}usl}
  \IEEEauthorblockA{%
     University of Applied Sciences Munich, Germany\\
    \{besim.shala, peter.mandl, andreas.humpe, martin.haeusl\}@hm.edu
  }
}

\maketitle

\emergencystretch=2em

\begin{abstract}
Online firestorms are rapid collective escalations of highly negative
user-generated content and may cause substantial reputational and economic
damage. Existing detectors usually work with volume signals, sentiment scores,
or predefined linguistic features. Such signals are useful, but they capture
contextual meaning shifts in evolving discussion threads only indirectly. This
paper proposes an LLM-based detection system with two operating modes. The first
mode classifies complete Reddit threads retrospectively by combining local
chunk-level assessments into a thread-level judgment. The second mode processes
threads sequentially and issues early warnings when a sliding window exceeds
calibrated thresholds. In this mode, the language model estimates three
firestorm indicators: negativity share, escalation level, and contributor count.
On a balanced Reddit dataset, the global mode achieves strong classification
performance, while the early warning mode reaches high recall and detects
escalating threads after only a small number of comments and distinct
contributors. The results indicate that LLMs can be used not only for static
judgment tasks, but also as repeated estimators in context-aware monitoring of
social media discourse.
\end{abstract}

\vspace{0.4em}


\section{Introduction}
\label{sec:intro}
\IEEEPARstart{O}{nline} firestorms, often referred to as ``shitstorms'' in
German language discourse, are sudden waves of highly negative user generated
content directed at organizations, brands, or individuals across social media
platforms \cite{pfeffer2014,rost2016}. They may escalate within hours and can
erode both reputation and public trust \cite{hansen2018}. Their consequences
are not limited to perception. Evidence from CSR related firestorms indicates a
moderate positive correlation between social media sentiment and stock market
performance \cite{dias2025}, which suggests that negative sentiment during such
events may also be associated with measurable downside risk. Since firestorms
usually rise and subside quickly, companies need to react before negative
sentiment gains further momentum \cite{qu2023}. Timely detection is therefore a
central precondition for crisis response, especially where short term brand
damage can translate into longer lasting reputational harm \cite{hansen2018}.

Existing detection approaches still rely heavily on aggregated volume signals
or predefined linguistic features computed at the post level \cite{koch2021}.
This is problematic because escalation is not only a property of isolated posts.
It emerges through relations between comments, changes in tone, and the
accumulation of mutually reinforcing contributions. Response oriented
frameworks have considered sequential dynamics at the firm level
\cite{herhausen2019}, but early detection itself is still commonly tied to
static feature thresholds rather than to the discourse sequence that precedes
escalation. Sentiment analysis is also fragile in this setting. Sarcasm, irony,
and implicit outrage can degrade classification accuracy, with documented case
studies reporting only 74\% accuracy and positive sentiment misclassification
rates above 90\% in highly sarcastic discourse \cite{nuortimo2020}. It remains
unclear how meaning shifts within coherent discussion threads can be converted
into actionable warning signals. This paper addresses that problem with a large
language model based detector that supports retrospective thread classification
and sequential early warning. Both modes use the model as a context sensitive
evaluator rather than as a keyword matcher or volume counter. The resulting
research question is:

\begin{center}
\vspace{-0.2em}
\emph{How can large language models be leveraged in a sequential, context aware
architecture to enable reliable, early detection of online firestorms in
social media discussion threads?}
\vspace{-0.4em}
\end{center}

The proposed system makes two contributions:

\begin{itemize}
    \setlength{\itemsep}{0.10em}
    \setlength{\parskip}{0pt}

    \item It translates large language model based evaluation into a
    deterministic and calibrated decision rule for firestorm monitoring. By
    using contextual model assessments rather than only keyword, sentiment, or
    volume signals, it supports warning decisions that can be integrated into
    real time monitoring workflows.

    \item It extends the LLM as a Judge paradigm \cite{gu2024} from static
    evaluation tasks to dynamic sequential monitoring. The model repeatedly
    estimates escalation signals over evolving discussion threads, linking
    firestorm theory to operational indicators such as negativity share,
    escalation level, and contributor count.
\end{itemize}

This provides a compact basis for evaluating LLM supported firestorm detection
under both retrospective and early warning conditions.

\section{Related Work}
\label{sec:related}

This section reviews three bodies of work that frame the present study:
conceptual definitions of online firestorms, prior detection and measurement
approaches, and early indicators that can be used for sequential warning.

\subsection{Defining and Characterizing Online Firestorms}

Online firestorms are characterized as sudden discharges of large quantities
of negative messages, typically manifesting as negative word-of-mouth and
complaint communication directed at a person, company, or group
\cite{pfeffer2014}. They represent crowd-based outrage, potentially
devastating storms of emotional and aggressive indignation in social media
\cite{rost2016}. Events of this kind are further marked by high message volume
and a distinctly negative, indignant opinion climate
\cite{pfeffer2014,johnen2018}. Participation is less the result of reflective
evaluation than of the cursory, reactive state characteristic of scrolling
through online communities, with the perception of being part of a collective
actor serving as the strongest driver of engagement \cite{gruber2020}.
Importantly, firestorms are primarily characterized by active attack strategies
such as public complaining, negative word-of-mouth, and aggressive commentary
rather than by silent boycott or patronage reduction \cite{delgado2019},
although heightened wrongness judgments toward firms can simultaneously foster
both vindictive complaining and patronage reduction \cite{chan2024}.

For the purposes of this study, the term \emph{online firestorm} refers to a
discussion thread in which collective outrage, negative condensation, and
escalating tonality are jointly observable, following the criteria established
in the foundational literature \cite{pfeffer2014,rost2016}. This operational
definition guides both dataset labeling and system design.

\subsection{Detection and Monitoring Approaches}

Prior work has translated firestorm indicators into several types of detection
and monitoring systems. An automated real-time firestorm detector combining
monitoring, sentiment analysis, and a statistical detection stage inspired by
epidemiological surveillance has been proposed, capturing activity in short time
intervals and comparing it against expected baselines to trigger an alarm when
deviations are statistically significant \cite{drasch2015}. Firestorm potential
in brand communities has been investigated through text mining, operationalizing
detection via high-arousal and low-arousal emotion indicators, linguistic style
matching, and tie strength between community members, yielding concrete
monitoring guidelines \cite{herhausen2019}. The early detection of
company-related firestorms on Twitter has been examined by systematically
evaluating feature variants that change significantly within the first hours,
outlining an hourly monitoring logic in which incoming posts are continuously
assessed for anomalous deviations \cite{koch2021}. Lexical change in negative
electronic word-of-mouth has been operationalized as an early indicator,
demonstrating that measurable linguistic shifts precede network-level escalation
signals such as retweets or mentions \cite{strathern2022}. A framework
distinguishing between trigger features observable at onset and evolving
firestorm features has further been proposed, providing both early warning
indicators and metrics for tracking ongoing escalation \cite{hansen2018}.

Beyond detection, researchers have also sought to measure and categorize the
severity of firestorms once they occur. Large-scale media analytics have been
used to establish a firestorm scale, showing that automated analysis can
efficiently process large volumes of social media data but typically reaches
accuracies of only between 70\% and 75\% \cite{nuortimo2020}. A standardized Social
Media Firestorm Scale has since been proposed, drawing an analogy to the
Saffir-Simpson hurricane scale and operationalizing firestorm severity along
three measurable dimensions: width (interaction volume), height (intensity of
negative sentiment), and length (duration of public discourse)
\cite{nuortimo2025}.

\subsection{Early Indicators of Firestorm Escalation}

The literature also points to several signals that occur early in firestorm
escalation. For system design, three recurring patterns are especially useful
because they are theoretically grounded and can be estimated from thread text.
They are not treated as a causal sequence, but as complementary warning signals
for the sequential mode developed in this paper.

The first is the degree of negativity in the discourse. A sharp increase in
negative comments within short time spans is considered a defining
characteristic of firestorm onset \cite{pfeffer2014}. The emerging opinion
climate is strongly negative and indignant, distinguishing firestorm threads
from ordinary critical discussions through the density and intensity of hostile
contributions rather than their mere presence \cite{johnen2018}.

The second indicator concerns linguistic and emotional escalation. Moral-emotional
language significantly amplifies the diffusion of content in social networks,
thereby accelerating collective outrage dynamics \cite{brady2017}. Early shifts
in language use, specifically a decline in self-referential phrasing and a rise
in negatively charged, emotionally loaded expressions, can signal an impending
firestorm before volume-level escalation becomes visible \cite{strathern2020}.

The third indicator is the number of actively contributing commenters.
Firestorms are characterized not only by negative tonality but by the
participation of many distinct users within a brief period \cite{johnen2018},
consistent with the piling-on dynamic that distinguishes collective outrage from
isolated negative commentary \cite{rost2016}.

The three indicators considered here are negativity share, escalation level,
and contributor count. Together, they inform the design of the sequential early
warning mode. Rather than claiming to identify a definitive or exhaustive set of
firestorm precursors, the goal is pragmatic: to select a small number of
theoretically motivated, estimable signals that together enable reliable early
detection in practice.

\subsection{Research Gap and Positioning}

This literature leaves a specific gap at the level of sequential thread
analysis. Existing approaches either depend on aggregated volume signals and
predefined linguistic features, which only approximate contextual meaning shifts,
or they are validated on narrow platform specific datasets with limited
transferability. To the best of our knowledge, context sensitive LLM evaluation
has not yet been examined as part of a sliding window architecture for
continuous thread level early warning. The present study therefore treats
firestorm detection as a sequence problem. Discussion threads are processed as
evolving discourse structures rather than as collections of independent
comments.

\section{Methodology}
\label{sec:method}

The methodology follows the unit of analysis used by the detector. It starts
from complete Reddit threads, assigns one global label per thread, and then
uses the same corpus to evaluate both retrospective recognition and early
warning.
\subsection{Reddit Dataset}

The detector requires labeled discussion threads rather than isolated posts.
Firestorm and non-firestorm cases therefore need to be represented in comparable
numbers, and each thread must retain its sequential structure, including comment
order and distinct contributors. The corpus also has to cover different topics
so that performance is not driven by artifacts of a single community.

The dataset consists of 200 publicly available Reddit discussion threads,
equally divided into 100 firestorm and 100 non-firestorm cases. Reddit was
chosen as the data source because the platform organizes discussions as coherent
comment threads within thematically bounded communities, i.e., subreddits,
enabling the observation of collective escalation dynamics within clearly defined
topical contexts \cite{qu2024}. As a text-centered platform, Reddit is
particularly suited for the early detection of firestorms, as collective outrage
dynamics can emerge through rapid and synchronous escalation of linguistic
content \cite{chan2025}. Empirical studies further confirm that Reddit user data
exhibits quality comparable to that of traditional research samples in terms of
reliability and validity \cite{jamnik2017}.

Data collection followed a passive approach, relying exclusively on already
existing, publicly accessible content without intervening in ongoing discussions
\cite{rochasilva2024}. Thread selection employed purposive sampling
\cite{campbell2020}, guided by established definitions of online firestorms
\cite{pfeffer2014,rost2016}. Subreddits were exploratively screened across a
spectrum of politically and thematically diverse communities to ensure that
clearly escalated threads, borderline cases, and uneventful discussions were
represented. Non-firestorm threads were drawn from predominantly neutral
subreddits characterized by fact-oriented discourse and clear discussion norms
(e.g., r/askscience, r/changemyview), serving as reference cases to assess the
detector's specificity.

Table~\ref{tab:subreddits} provides an overview of the subreddit distribution
across both classes, grouped by thematic category.

\begin{table}[t]
\caption{\textbf{Subreddit Distribution by Thematic Category}}
\label{tab:subreddits}
\vspace{3pt}
\centering
\small
\setlength{\tabcolsep}{0pt}
\renewcommand{\arraystretch}{1.08}
\begin{tabular}{@{}p{2.05cm}r@{\hspace{0.30cm}}r@{\hspace{0.45cm}}p{3.85cm}@{}}
\hline
\textbf{Category} & \textbf{FS} & \textbf{NFS} & \textbf{Example Subreddits} \\
\hline
Politics      & 40  & 3   & conservative, socialism \\
Consumer      & 15  & 0   & anticonsumption, fuckamazon \\
Entertainment & 14  & 0   & popculture, fauxmoi \\
News          & 12  & 7   & worldnews, nottheonion \\
Science       & 0   & 85  & askscience, askacademia \\
Meta          & 9   & 5   & subredditdrama, changemyview \\
Gaming        & 10  & 0   & chess, codwarzone \\
\hline
\textbf{Total} & \textbf{100} & \textbf{100} & \\
\hline
\noalign{\vskip 4pt}
\multicolumn{4}{@{}p{7.00cm}@{}}{\footnotesize
FS = firestorm; NFS = non-firestorm threads.}
\end{tabular}
\end{table}

The asymmetric distribution of subreddits across classes reflects the
deliberate sampling strategy. Non-firestorm threads were drawn from a small
number of communities characterized by consistently neutral discourse norms and
fact-oriented discussion, providing a stable reference baseline. Firestorm
threads, by contrast, were sampled across a broader spectrum of topical contexts
to ensure that the detector is evaluated against diverse escalation patterns
rather than community-specific artifacts.

Table~\ref{tab:stats} summarizes the structural properties of the dataset.
Firestorm threads exhibit substantially higher participation volume, with nearly
twice as many comments and unique contributors on average, while individual
comments are markedly shorter on average ($M=146.2$, $SD=61.8$ vs.\
$M=328.1$, $SD=117.6$ characters). Here, $M$ denotes the arithmetic mean and
$SD$ denotes the standard deviation.

\begin{table}[t]
\caption{\textbf{Descriptive Statistics by Class}}
\label{tab:stats}
\vspace{3pt}
\centering
\small
\setlength{\tabcolsep}{2pt}
\renewcommand{\arraystretch}{1.08}
\begin{tabular}{L{2.25cm}cc}
\hline
Metric & FS ($n{=}100$) & NFS ($n{=}100$) \\
\hline
Avg.\ comments     & 176.7 ($\pm$129.3) & 96.6 ($\pm$73.3) \\
Med.\ comments     & 136.0              & 80.5             \\
Avg.\ contributors & 123.8 ($\pm$92.7)  & 66.4 ($\pm$54.0) \\
Med.\ contributors & 97.0               & 57.5             \\
Avg.\ length (ch)  & 146.2 ($\pm$61.8)  & 328.1 ($\pm$117.6) \\
Med.\ length (ch)  & 131.4              & 314.2            \\
\hline
\noalign{\vskip 4pt}
\multicolumn{3}{L{7.6cm}}{\footnotesize
FS = firestorm; NFS = non-firestorm; Avg. = arithmetic mean ($M$);
Med. = median; $SD$ = standard deviation. Values in parentheses report
standard deviations.}
\end{tabular}
\end{table}

Each thread was labeled with a single, global binary annotation, distinguishing
firestorm from non-firestorm cases at the thread level. The labeling followed
the definitional criteria of collective outrage, negative condensation, and
escalating tonality \cite{pfeffer2014,rost2016}. An explicit annotation of the
exact escalation onset was deliberately omitted, as such a point is empirically
ambiguous and highly dependent on interpretive choices. This labeling strategy
is consistent with principles of reliable content coding, which require that
categorical decisions rest on clearly formulated criteria applied consistently
\cite{artstein2008}.

\subsection{System Architecture}

The detector has two modes with the same thread level reference label but
different decision points. The \emph{Global Recognition Mode} uses the complete
thread and provides a retrospective performance baseline. The \emph{Early
Warning Mode} processes a thread while it grows and asks when a warning can be
issued. Numerical parameters were selected through iterative empirical testing;
they are practical optima for this setup and are not claimed to be globally
optimal. The system prompts were refined during development and are reported in
the Appendix.

\subsection{Global Recognition Mode}

The global mode treats the Reddit thread as a complete document and returns one
binary thread level classification. Because discussion threads
frequently exceed the practical context limits of current Transformer models,
whose self-attention mechanism scales quadratically in compute and memory with
sequence length \cite{qin2022}, the input is segmented into chunks up to
12,000 characters each. This segmentation proceeds comment by comment in
chronological order, preserving local discourse coherence within each chunk
while ensuring full coverage of the thread. Research on chunk-based
representations confirms that segmentation can preserve essential semantic
content while reducing input size \cite{li2024}.

\begin{figure}[t!]
  \centering
  \includegraphics[width=\columnwidth]{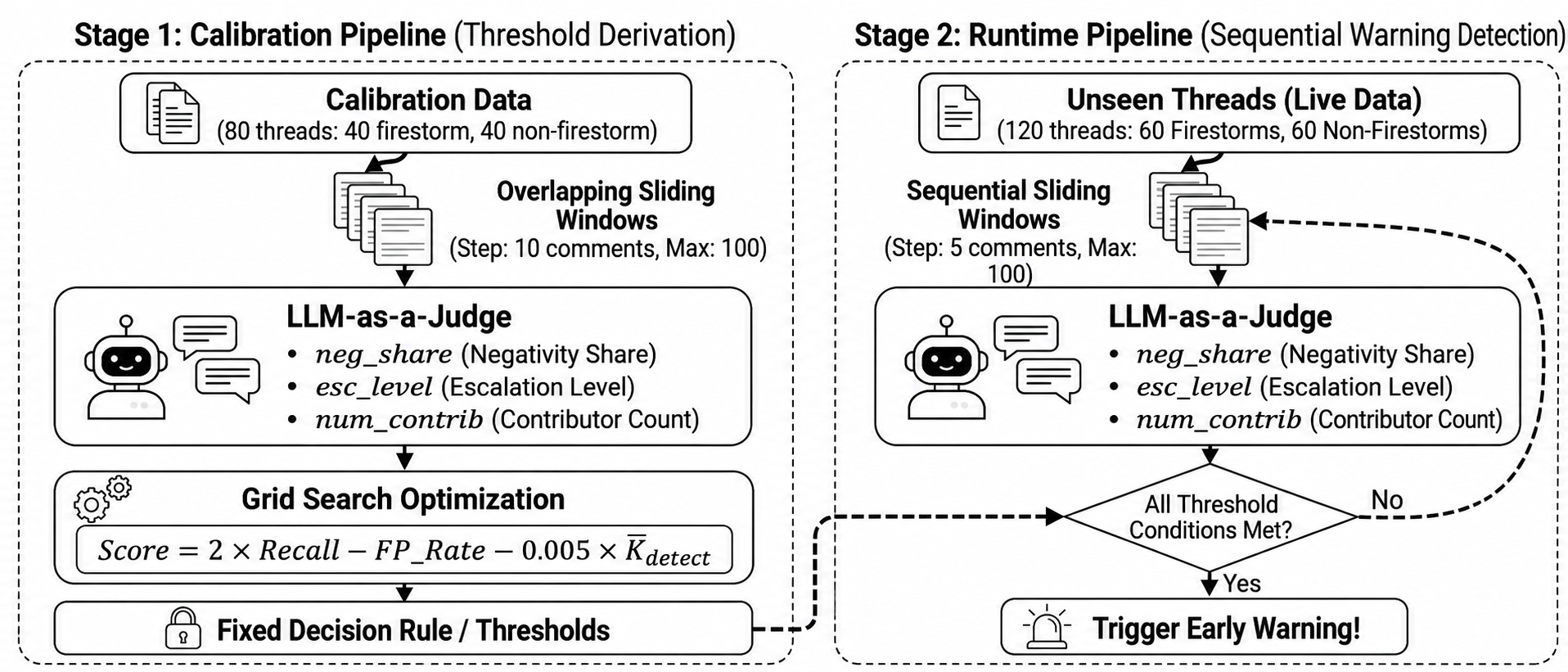}
  \caption{\textbf{Global Recognition Mode pipeline with local LLM-based chunk
    assessment (Stage~1) and global thread-level classification (Stage~2).}}
  \label{fig:global}
\end{figure}

The overall architecture is illustrated in Fig.~\ref{fig:global}.

Each chunk is then independently assessed by a large language model (LLM)
operating under a fixed system prompt that encodes the firestorm definition and
assessment criteria \cite{pfeffer2014,rost2016} (Stage~1 in Fig.~\ref{fig:global}).
The prompt-based assessment follows the paradigm of prompt-based learning, in
which inputs are transformed into a prompt template with defined slots and the
model output is mapped to the desired labels via a fixed answer space
\cite{liu2021}. For each chunk, the LLM produces two outputs: a local binary
judgment (firestorm / no-firestorm) and a concise summary capturing the
conflict-relevant content of the segment.

In the second stage (Fig.~\ref{fig:global}, right panel), all chunk summaries
and their associated local labels are concatenated in chronological order into
a compressed meta-description of the entire thread. This condensed
representation is then submitted to the LLM in a final call that produces the
global thread-level classification together with a brief justification. This
two-stage architecture follows the principle of hierarchical context merging,
in which local segment representations are progressively integrated into a
global representation \cite{ou2025}. Song et al.\ \cite{song2024} demonstrate
with HOMER that such hierarchical merging of chunk-level information improves
long-context task performance compared to baselines, supporting the practical
viability of this approach.

\subsection{Sequential Early Warning Mode}

The early warning mode adds a temporal perspective. It does not wait for the
thread to be complete, but evaluates the growing discussion and triggers a
warning once the calibrated conditions are met.

\begin{figure}[t!]
  \centering
  \includegraphics[width=\columnwidth]{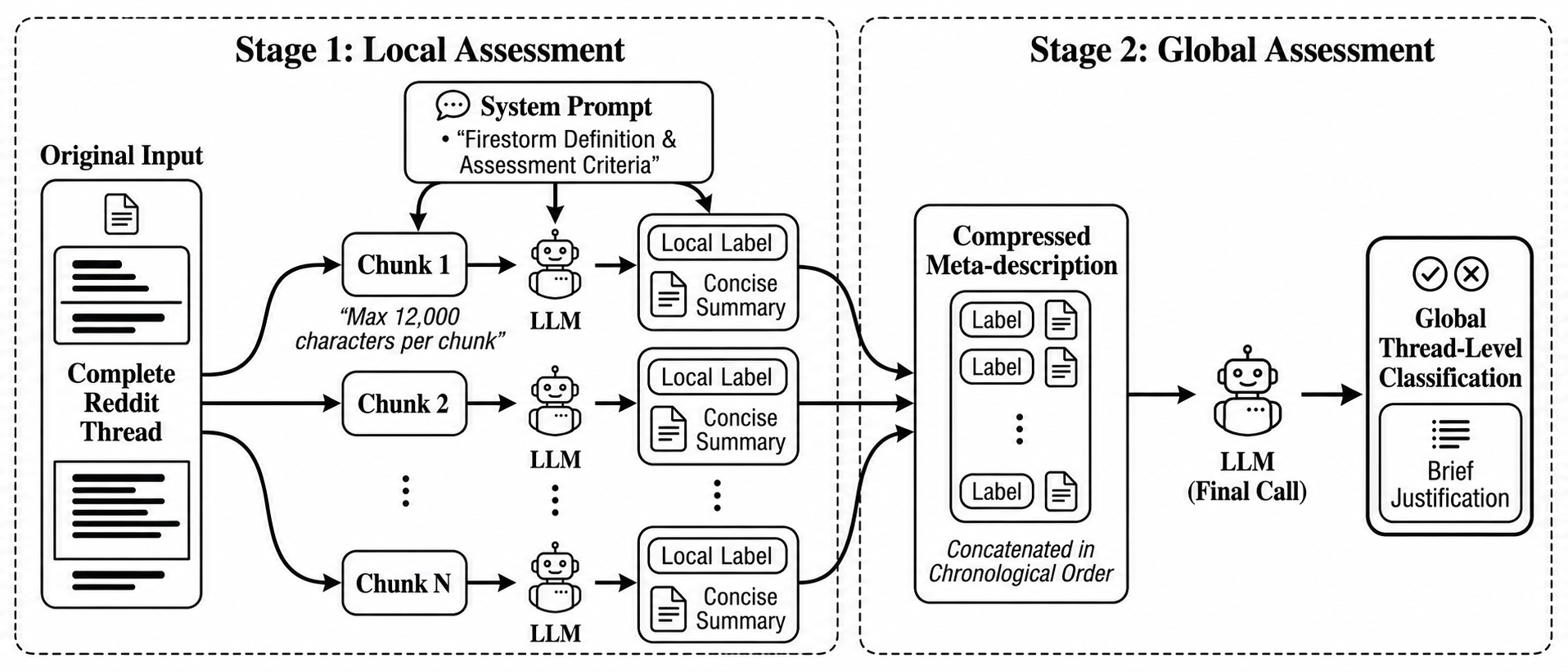}
  \caption{\textbf{Two-stage Early Warning Mode with threshold calibration
    (Stage~1) and sequential warning detection via sliding windows (Stage~2).}}
  \label{fig:earlywarning}
\end{figure}

The architecture is divided into two stages: a calibration pipeline and a
runtime pipeline, as illustrated in Fig.~\ref{fig:earlywarning}.

In the calibration pipeline (Stage~1 of Fig.~\ref{fig:earlywarning}), a
balanced subsample of 80 threads (40 firestorms, 40 non-firestorms) was used
to empirically derive a fixed decision rule. For each thread in this subsample,
overlapping sliding windows of increasing size are formed in fixed steps of 10
comments, up to a maximum of 100 comments. This approach draws on the
overlapping chunking principle established in the Retrieval-Augmented Generation
literature, where consecutive text segments share content at their boundaries
to preserve contextual continuity across the analysis \cite{gao2024}. For each
window, the LLM estimates three early indicators in the role of a standardized
judge (LLM-as-a-Judge), following the paradigm in which LLMs produce structured
numerical scores within a prompted evaluation context \cite{gu2024}. The three
indicators are grounded in the communication science literature on firestorm
precursors (see Section~\ref{sec:related}):

\begin{itemize}
  \item \emph{Negativity share} (\texttt{neg\_share}): the estimated proportion
        of negative comments in the current window.
  \item \emph{Escalation level} (\texttt{esc\_level}): the degree of
        reinforcement, piling-on, and intensification.
  \item \emph{Contributor count} (\texttt{num\_contrib}): the number of distinct
        users whose comments are classified as conflict-amplifying within the
        window.
\end{itemize}

\subsubsection{Threshold Calibration via Grid Search}

Based on the calibration data, a grid search \cite{bergstra2012} is performed
over candidate threshold combinations for all three indicators plus a positive
status signal (emerging or full firestorm). The grid search scanned the
following candidate values:
\begin{equation}
\label{eq:grid_candidates}
\begin{aligned}
\mathtt{neg\_share}   &\in \{0.30, 0.40, 0.50, 0.60, 0.70\},\\
\mathtt{esc\_level}   &\in \{0.20, 0.30, 0.40, 0.50, 0.60\},\\
\mathtt{min\_contrib} &\in \{3, 5, 8, 12, 15\}.
\end{aligned}
\end{equation}
The grid therefore contains 125 candidate threshold combinations. The selected
combination maximizes the following scoring function:
\begin{equation}
\label{eq:score}
\mathrm{score}=2\,\mathrm{Rec}_{\mathrm{FS}}-\mathrm{FPR}-\lambda\bar{c}_{\mathrm{det}},
\end{equation}
where $\mathrm{Rec}_{\mathrm{FS}}$ is the recall on the firestorm class,
$\mathrm{FPR}$ is the false-positive rate, $\bar{c}_{\mathrm{det}}$ is the mean
number of comments until detection, and $\lambda$ is a mild penalty weight.
This explicit prioritization of recall reflects the asymmetric cost structure
of firestorm detection, where missing an escalation carries greater risk than
triggering a false alarm \cite{herhausen2019}. The choice of thresholds on a
separate calibration set follows the standard two-stage approach of first
producing continuous scores and then selecting a threshold that maximizes a
target metric on validation data \cite{nan2012}.

Once the thresholds are fixed, the runtime pipeline (Stage~2 of
Fig.~\ref{fig:earlywarning}) applies them unchanged to previously unseen
threads. Each thread is processed in sliding windows of 5 comments per step,
up to a maximum of 100 comments. At each window position, the LLM estimates
the three indicators, and the fixed decision rule is evaluated. A warning is
triggered as soon as all threshold conditions are jointly met; otherwise, the
analysis continues to the next window until either a warning is issued or the
thread ends. Algorithm~\ref{alg:ew_runtime} formalizes this procedure.

\begin{algorithm}[t]
\caption{Early Warning Runtime Pipeline}
\label{alg:ew_runtime}
\footnotesize
\begin{algorithmic}[1]
\Require Thread $T=(c_i)_{i=1}^N$, thresholds $\theta=(\tau_n,\tau_e,\tau_k)$
\Require step size $s=5$, maximum horizon $H=100$
\Ensure warning flag $\mathit{warn}$ and detection point $d$
\State $\mathcal{F}\gets\{\text{emerging},\text{full}\}$
       \Comment{firestorm statuses}
\State $\mathit{warn}\gets\mathrm{false}$; $d\gets\bot$; $\mathit{pos}\gets s$
\While{$\mathit{pos}\leq\min(N,H)$ \textbf{and} $\neg\mathit{warn}$}
  \State $W\gets(c_1,\ldots,c_{\mathit{pos}})$
         \Comment{current prefix window}
  \State $(n,e,k,q)\gets\mathrm{LLMJudge}(W)$
         \Comment{estimate indicators}
  \State $\mathit{negOK}\gets(n\geq\tau_n)$;
         $\mathit{escOK}\gets(e\geq\tau_e)$
  \State $\mathit{cntOK}\gets(k\geq\tau_k)$;
         $\mathit{stateOK}\gets(q\in\mathcal{F})$
  \If{$\mathit{negOK}\land\mathit{escOK}\land
       \mathit{cntOK}\land\mathit{stateOK}$}
    \State $\mathit{warn}\gets\mathrm{true}$;
           $d\gets\mathit{pos}$
           \Comment{issue warning}
  \EndIf
  \State $\mathit{pos}\gets\mathit{pos}+s$
         \Comment{advance window}
\EndWhile
\State \Return $\mathit{warn}$, $d$
\end{algorithmic}

\vspace{2pt}
\footnotesize
\textit{Note.} $n$ denotes negative share, $e$ escalation level, $k$ contributor count,
and $q$ thread status. The value $d=\bot$ indicates that no warning was triggered.
\end{algorithm}

\section{Experiments}
\label{sec:experiments}

The experiments test the system in the two settings for which it is designed.
The first setting asks whether complete Reddit threads can be classified
reliably after the fact. The second asks whether a warning can be issued early
during sequential processing. Evaluation therefore covers classification
performance, API call complexity, and detection timeliness.

\subsection{Experimental Setup}

All experiments use OpenAI's GPT-4o mini as the underlying language model. The
model was chosen for pragmatic reasons: the detector requires many repeated
model calls, and GPT-4o mini was designed for cost efficient applications that
chain or parallelize calls at scale \cite{openai2024}. Models of this class can
perform complex assessment tasks from instructions alone and do not require
task specific fine tuning for the prompt based evaluation design used here
\cite{openai2023}.

Model behavior is controlled through dedicated system prompts that define the
assessment criteria and enforce structured JSON output via schema constraints,
ensuring that responses can be parsed automatically without manual intervention.
In the global mode, the system prompt encodes the firestorm definition and
specifies the output format for both classification stages. In the early warning
mode, a separate system prompt instructs the model to estimate the three
numerical indicators together with a categorical status assessment.

Table~\ref{tab:apicalls} summarizes the number of LLM API calls required per
thread in each operating mode.

\begin{table}[t]
\caption{\textbf{API Call Complexity per Thread}}
\label{tab:apicalls}
\vspace{3pt}
\centering
\footnotesize
\setlength{\tabcolsep}{2pt}
\renewcommand{\arraystretch}{1.08}
\begin{tabular}{L{1.35cm}L{2.85cm}L{3.00cm}}
\hline
Mode & Stage & Calls per thread \\
\hline
Global        & Chunk assessment             & $\lceil L/C \rceil$ \\
Global        & \mbox{Thread classification} & $1$ \\
Global        & Total                        & $\lceil L/C \rceil + 1$ \\
Early Warning & Calibration                  & $\lfloor \min(N,100)/s_{\mathrm{cal}} \rfloor$ \\
Early Warning & Runtime                      & $1$ to $\lfloor \min(N,100)/s_{\mathrm{run}} \rfloor$ \\
\hline
\noalign{\vskip 4pt}
\multicolumn{3}{L{7.25cm}}{\footnotesize
$L$ = thread length in characters; $C$ = chunk size (12{,}000 characters);
$N$ = number of comments; $s_{\mathrm{cal}}$ = calibration step size (10);
$s_{\mathrm{run}}$ = runtime step size (5). Runtime calls terminate early upon
first threshold breach.}
\end{tabular}
\end{table}

In the global mode, computational cost scales linearly with thread length, as
each chunk requires one independent assessment call plus one final aggregation
call. In the early warning mode, the number of calls depends on both the thread
length and whether the decision rule triggers early. In the best case, a
firestorm is detected after a single window evaluation; in the worst case, the
full sequence of windows up to the 100-comment horizon is processed without
triggering.

\subsection{Evaluation Metrics}

The evaluation is anchored at the thread level for both modes, with the global
binary label (firestorm vs.\ non-firestorm) serving as ground truth. For the
global mode, the system's final thread-level decision is compared against the
ground truth label via the standard confusion matrix (TP, FP, TN, FN).
Precision, Recall, and F1-score are computed per class. Accuracy is reported as
a summary measure. Because the test set is balanced (100 threads per class),
accuracy is not inflated by a dominant majority class, which makes it a
meaningful aggregate in this setting \cite{chicco2020}. Since the global mode
does not involve a calibration step, all 200 threads are used for evaluation.

For the early warning mode, evaluation is performed exclusively on the 120
threads (60 firestorm, 60 non-firestorm) that were not used during threshold
calibration, ensuring a strict separation between parameter tuning and
performance assessment. In addition to standard classification metrics, two
temporal measures quantify detection timeliness:

\begin{itemize}
  \item \emph{Mean comments until detection}: the mean number of comments seen
        before the first warning is triggered across all correctly detected
        firestorm threads.
  \item \emph{Mean contributing users until detection}: the mean number of
        distinct contributing users at the point of first detection.
\end{itemize}

The false-positive rate (FPR) is reported separately as a dedicated measure of
alarm burden. Together with Recall on the firestorm class, the FPR enables a
pragmatic assessment of the trade-off between early sensitivity and alarm
reliability.

\section{Results}
\label{sec:results}

The results are reported separately for global thread classification and early
warning. Section~\ref{sec:discussion} then discusses how these findings should
be interpreted.

\subsection{Global Classification}

Table~\ref{tab:global_results} summarizes the classification performance and
confusion matrix for the global recognition mode, evaluated on all 200 threads.

\begin{table}[t]
\caption{\textbf{Global Mode: Classification Metrics and Confusion Matrix}}
\label{tab:global_results}
\vspace{3pt}
\centering
\footnotesize
\setlength{\tabcolsep}{2pt}
\renewcommand{\arraystretch}{1.08}
\begin{tabular}{L{2.05cm}C{1.15cm}C{1.15cm}C{1.05cm}C{1.15cm}}
\hline
\multicolumn{5}{l}{\textit{(a) Confusion Matrix}} \\
\hline
             & \multicolumn{2}{c}{Pred.\ FS} & \multicolumn{2}{c}{Pred.\ NFS} \\
\hline
True FS      & \multicolumn{2}{c}{88~(TP)}   & \multicolumn{2}{c}{12~(FN)} \\
True NFS     & \multicolumn{2}{c}{5~(FP)}    & \multicolumn{2}{c}{95~(TN)} \\
\hline
\noalign{\vskip 7pt}
\multicolumn{5}{l}{\textit{(b) Per-Class Metrics}} \\
\hline
Class        & Prec. & Rec. & F1   & Supp. \\
\hline
Firestorm    & 0.95  & 0.88 & 0.91 & 100 \\
NFS          & 0.89  & 0.95 & 0.92 & 100 \\
\hline
\noalign{\vskip 4pt}
\multicolumn{5}{L{7.35cm}}{\footnotesize
FS = firestorm; NFS = non-firestorm. Overall accuracy: 0.915.}
\end{tabular}
\end{table}

Table~\ref{tab:global_results} reports both the confusion matrix and the
per-class metrics. The global mode reaches an overall accuracy of 0.915. For
the firestorm class, Precision is 0.95 and Recall is 0.88, which means that the
system rarely assigns the firestorm label incorrectly but misses 12 of the 100
firestorm threads. The resulting F1-score is 0.91. For non-firestorm threads,
Recall is higher at 0.95, while Precision is 0.89. The confusion matrix contains
88 true positives, 95 true negatives, 5 false positives, and 12 false negatives.
This pattern shows that the global mode is somewhat conservative when assigning
the firestorm label.

\subsection{Early Warning}

Tables~\ref{tab:ew_metrics} to~\ref{tab:ew_temporal} present the classification
metrics, confusion matrix, and temporal detection results for the early warning
mode. The calibrated thresholds are reported in the text rather than in a
separate table: $\texttt{min\_neg}=0.4$, $\texttt{min\_esc}=0.2$, and
$\texttt{min\_contrib}=3$.

\begin{table}[t]
\caption{\textbf{Early Warning Mode: Per-Class Classification Metrics}}
\label{tab:ew_metrics}
\vspace{3pt}
\centering
\footnotesize
\setlength{\tabcolsep}{4pt}
\renewcommand{\arraystretch}{1.08}
\begin{tabular}{L{2.15cm}C{1.0cm}C{1.0cm}C{1.0cm}C{1.25cm}}
\hline
Class & Prec. & Rec. & F1 & Support \\
\hline
Firestorm    & 0.82 & 0.98 & 0.89 & 60 \\
No-Firestorm & 0.98 & 0.78 & 0.87 & 60 \\
\hline
\end{tabular}
\end{table}

\begin{table}[t]
\caption{\textbf{Early Warning Mode: Confusion Matrix}}
\label{tab:ew_cm}
\vspace{3pt}
\centering
\footnotesize
\setlength{\tabcolsep}{7pt}
\renewcommand{\arraystretch}{1.08}
\begin{tabular}{L{2.15cm}C{1.55cm}C{1.55cm}}
\hline
                   & Pred.\ FS & Pred.\ NFS \\
\hline
True FS            & 59 (TP) &  1 (FN) \\
True NFS           & 13 (FP) & 47 (TN) \\
\hline
\noalign{\vskip 4pt}
\multicolumn{3}{L{6.60cm}}{\footnotesize
FS = firestorm; NFS = non-firestorm.}
\end{tabular}
\end{table}

\begin{table}[t]
\caption{\textbf{Early Warning Mode: Temporal Detection Metrics}}
\label{tab:ew_temporal}
\vspace{3pt}
\centering
\footnotesize
\setlength{\tabcolsep}{5pt}
\renewcommand{\arraystretch}{1.08}
\begin{tabular}{L{5.10cm}C{1.25cm}}
\hline
Metric & Value \\
\hline
Mean comments until detection & 8.56 \\
Mean comment ratio (det./thread avg.) & 4.8\% \\
Mean contributing users until detection & 4.02 \\
\hline
\noalign{\vskip 4pt}
\multicolumn{2}{L{6.75cm}}{\footnotesize
Comment ratio = 8.56 / 176.7 (mean firestorm thread length).}
\end{tabular}
\end{table}

The calibrated thresholds set the minimum negative share to 0.4, the minimum
escalation level to 0.2, and the minimum contributor count to 3.
Table~\ref{tab:ew_metrics} shows that Recall for the firestorm class reaches
0.98, meaning 59 out of 60 firestorm threads are correctly detected. Precision
for the firestorm class is 0.82, reflecting the deliberate sensitivity bias
introduced by the scoring function. The non-firestorm class exhibits the
inverse pattern: Precision of 0.98 at Recall of 0.78. Table~\ref{tab:ew_cm}
shows 59 true positives and 47 true negatives, with 13 false positives and only
1 false negative. The resulting $\text{FPR}=13/60\approx0.217$, meaning
approximately 22\% of non-firestorm threads trigger a warning.
Table~\ref{tab:ew_temporal} reports that on average the system triggers its
first warning after 8.56 comments, involving a mean of 4.02 distinct
contributing users at the point of detection. To contextualize this figure, the
mean firestorm thread in the test set contains 176.7 comments; detection at
8.56 comments thus occurs after approximately 4.8\% of a thread's eventual
volume has materialized, indicating that the early warning mode responds well
before the escalation pattern has fully developed.

\section{Discussion and Limitations}
\label{sec:discussion}

This section discusses what the empirical results imply for firestorm detection,
how they compare with prior approaches, and where the current validation remains
limited.

\subsection{Discussion}

The results are encouraging in both operating modes, but they require a careful
reading because the task combines classification, temporal detection, and
operational risk.

In the global mode, the hierarchical LLM-based classifier clearly exceeds the
accuracy range between 70\% and 75\% reported for traditional automated
sentiment analysis in firestorm monitoring \cite{nuortimo2020}. The improvement
is plausibly linked to two architectural choices. Chunk-level assessment
preserves local discourse context that would be lost in comment-by-comment
classification, a limitation also noted in contextual hate speech detection
\cite{perez2023}. The subsequent merging step then combines these local signals
into a thread-level judgment, in line with evidence that hierarchical merging
can improve long-context performance \cite{song2024}. This design makes it
possible to represent piling-on behavior and tonal shifts across comment
sequences, which dictionary-based and feature-engineered approaches
\cite{herhausen2019,koch2021} capture only indirectly.

The early warning mode addresses the harder operational question of timeliness.
Its firestorm Recall of 0.98 shows that almost all escalating threads are
flagged. The cost is a higher false-positive rate, but this trade-off is
expected for early warning systems. Even under idealized conditions, early
warning indicators involve a pronounced balance between sensitivity and false
alarms \cite{boettiger2012}. In the firestorm setting, this sensitivity bias is
reasonable when the cost of reviewing a benign thread is lower than the
potential reputational and financial cost of a missed escalation
\cite{herhausen2019}.

\subsubsection{Structural Filtering and Detection Timeliness}

The contributor count threshold (\texttt{min\_contrib}~=~3) operationalizes the
multi-actor criterion from firestorm theory \cite{johnen2018}. A warning is not
triggered by negativity alone; it also requires evidence of collective dynamics.
This adds a structural filter that pure sentiment assessment lacks.

The temporal detection metrics further support the practical value of the
approach. Detection after an average of 8.56 comments and 4.02 distinct
contributing users indicates that the system responds to initial escalation
signals well before broad participation has materialized. This aligns with the
theoretical observation that online firestorms are triggered by a sudden,
targeted collective brawl characterized by an immediate shift towards aggressive
language against a single entity \cite{strathern2020}. Identifying the threat
before broad participation is observable allows companies to consciously deploy
the most effective immediate mitigation strategy \cite{qu2023}.

The comparison with prior work is therefore mainly conceptual. Volume-based
detectors identify deviations from activity baselines \cite{drasch2015}, while
the present approach evaluates the content of the emerging discourse. Feature
engineered approaches depend on predefined linguistic indicators
\cite{herhausen2019}; the LLM-based evaluation can also use implicit meaning,
irony, and local context when judging whether a thread is escalating.

\subsection{Theoretical and Practical Contributions}

At the conceptual level, the study extends the LLM-as-a-Judge paradigm from
static evaluation to sequential monitoring. The model is not asked to score a
finished artifact once. Instead, it repeatedly estimates indicators over an
evolving discussion and feeds these estimates into a deterministic decision
rule. This connects qualitative firestorm theory with reproducible operational
signals, namely negativity share, escalation level, and contributor count.

For organizations, the two modes map to different monitoring tasks. The global
recognition mode can be used retrospectively to screen thread archives and
identify past escalation events. The early warning mode is closer to an
operational workflow: a warning can surface a thread for human review, support a
short situation summary from the most recent window, and notify roles such as
communications, moderation, or legal teams. Since thresholds are calibrated
explicitly, the same procedure can be repeated for organization specific data,
community norms, or different risk tolerances.

\subsection{Limitations}
Several limitations are relevant when interpreting the results. They concern
the data source, the annotation procedure, temporal validation, model choice,
and the range of signals available to the detector.

\textbf{Platform dependence.} The dataset consists exclusively of Reddit
discussion threads. Reddit-specific characteristics, including community norms,
moderation practices, and visibility mechanisms, shape the observable discourse
patterns \cite{proferes2021}. The extent to which the observed detection
performance generalizes to platforms with different interaction structures such
as Twitter/X or Facebook remains an open question.

\textbf{Single-annotator bias.} The ground-truth labels were assigned by a
single annotator following the definitional criteria of collective outrage,
negative condensation, and escalating tonality \cite{pfeffer2014,rost2016}.
While this is consistent with principles of reliable content coding
\cite{artstein2008}, the absence of independent second annotation means that
systematic labeling biases cannot be fully ruled out.

\textbf{Absence of escalation onset annotation.} The deliberate omission of an
exact escalation start point means that the early warning mode cannot be
validated against a precise temporal ground truth. The reported temporal
metrics provide a pragmatic operationalization of detection timeliness but are
approximations rather than precise measurements relative to the true escalation
moment.

\textbf{Model dependence.} The system relies on a specific LLM (GPT-4o mini).
The stability of results across model versions, alternative providers, and
prompt variations has not been systematically evaluated. Since prompt-based
assessments are sensitive to instruction formulation \cite{zhou2022}, the exact
threshold configuration may not transfer directly to other model variants.
However, the pipeline architecture is model-agnostic: the calibration procedure
can be re-applied to any model that produces structured numerical outputs.

\textbf{Reliability of LLM indicator estimates.} The early warning mode relies
on a general-purpose LLM to estimate three numerical indicators
(\texttt{neg\_share},\allowbreak{} \texttt{esc\_level},\allowbreak{} \texttt{num\_contrib}) from
unstructured text. While structured JSON output constraints and low-temperature
sampling reduce variance, the accuracy of these estimates has not been validated
against an independent ground truth at the comment level. In particular, LLMs
are susceptible to hallucination and instruction-following inconsistencies
\cite{zhou2022}, which may cause systematic over- or underestimation of
individual indicators. The observed calibration results provide only indirect evidence of sufficient
estimability. They show that calibrated thresholds perform well on held-out data,
but they do not prove that each individual indicator is estimated accurately. A
direct comparison between LLM-estimated indicator values and human-annotated
ground truth would strengthen confidence in the approach.

\textbf{Text-only focus.} The proposed approach relies exclusively on textual
content and does not incorporate author-level or network-level signals such as
social capital, follower reach, or tie strength between community members. Prior
research suggests that such structural features can significantly influence the
diffusion and escalation of firestorm content \cite{jontgen2020}.

\section{Conclusion}
\label{sec:conclusion}

This study evaluated an LLM-based firestorm detector on a balanced dataset of
200 Reddit threads. The main finding is that thread structure matters. When
comments are treated as an evolving discourse rather than as isolated posts, the
system can capture piling-on dynamics, tonal shifts, and implicit escalation
patterns that are difficult to represent with predefined features or volume
baselines alone.

The global recognition mode achieves an F1-score of 0.91 and an accuracy of
0.915, which is above the accuracy range between 70\% and 75\% reported for
conventional automated approaches \cite{nuortimo2020}. The early warning mode
detects 98\% of firestorms after an average of 8.56 comments. This indicates
that theoretically grounded early indicators can be estimated by a
general-purpose LLM well enough to support calibrated and deterministic early
warning. The findings therefore extend the LLM-as-a-Judge paradigm to a
sequential monitoring context and provide a reproducible operationalization of
qualitative firestorm theory.

Future work should first test generalizability across further platforms and
language contexts \cite{proferes2021}. A systematic comparison across multiple
LLMs would also clarify how sensitive the results are to model choice and prompt
formulation \cite{zhou2022}. The temporal evaluation could be strengthened with
more fine-grained escalation onset annotations. The decision rule could further
be extended with uncertainty estimates, adaptive window sizes, or case based
threshold calibration. Finally, the system should be studied in operational
incident workflows, where warnings are routed to communications or moderation
teams. In such settings, context sensitive LLM evaluation combined with explicit
calibration offers a practical path toward earlier detection of collective
escalation.

\section*{Acknowledgment}
This work was supported in part by the Bavarian State Ministry of Economic
Affairs, Regional Development and Energy under the BayVFP funding line
Digitalization, program area Information and Communication Technology, through
the project ``NetiChecker'' under grant number DIK-2507-0017//DIK0787/01.

\section*{Disclaimer}

Large Language Models were used as supporting tools for drafting selected text
passages and for creating and revising figures. The authors retain full and
sole responsibility for the content, cited sources, argumentation, and final
version of this article.

\bibliographystyle{IEEEtran}
\bibliography{refs}

\end{document}